\documentclass{article}

\usepackage[accepted]{icml2019}

\usepackage[utf8]{inputenc} 
\usepackage[T1]{fontenc}    
\usepackage{hyperref}       
\usepackage{url}            
\usepackage{booktabs}       
\usepackage{amsfonts}       
\usepackage{nicefrac}       
\usepackage{microtype}      
\usepackage{float}
\usepackage{amsmath}

\usepackage{graphicx}
\graphicspath{ {.} }

\icmltitlerunning{Dropout is a special case of the stochastic delta rule: faster and more accurate deep learning}

\begin{document}

\twocolumn[
\icmltitle{Dropout is a special case of the stochastic delta rule: faster and more accurate deep learning}

\icmlsetsymbol{equal}{*}

\begin{icmlauthorlist}
\icmlauthor{Noah Frazier-Logue}{equal,ru}
\icmlauthor{Stephen José Hanson}{equal,ru}
\end{icmlauthorlist}

\icmlaffiliation{ru}{Rutgers University Brain Imaging Center, Rutgers University Newark, Newark, New Jersey}

\icmlcorrespondingauthor{Stephen José Hanson}{jose@rubic.rutgers.edu}
\icmlcorrespondingauthor{Noah Frazier-Logue}{n.frazier.logue@nyu.edu}

\icmlkeywords{Machine Learning, ICML, Dropout, SDR, Deep Learning}

\vskip 0.3in
]

\printAffiliationsAndNotice{\icmlEqualContribution}

\begin{abstract}

Multi-layer neural networks have lead to remarkable performance on many kinds of benchmark tasks in text, speech and image processing.  Nonlinear parameter estimation in hierarchical models is known to be subject to overfitting and misspecification.   One approach to these estimation and related problems (local minima, colinearity, feature discovery etc.)  is called Dropout (Hinton, et al 2012, Baldi et al 2016). The Dropout algorithm removes hidden units according to a Bernoulli random variable with probability $p$ prior to each  update, creating random "shocks" to the network that are  averaged over updates.  In this paper we will show that Dropout is a special case of a more general model published originally in 1990 called the Stochastic Delta Rule, or SDR (Hanson, 1990).  SDR redefines each weight in the network as a random variable with mean $\mu_{w_{ij}}$ and standard deviation $\sigma_{w_{ij}}$.  Each weight random variable is sampled on each forward activation, consequently creating an exponential number of potential networks with shared weights.    Both parameters are updated according to prediction error, thus resulting in weight noise injections that reflect a local history of prediction error and local model averaging.  SDR therefore implements a more sensitive local gradient-dependent simulated annealing per weight converging in the limit to a Bayes optimal network.   Tests on standard benchmarks (CIFAR) using a modified version of DenseNet shows the SDR outperforms standard Dropout in test error by approx. $17\%$ with DenseNet-BC 250 on CIFAR-100 and approx. $12-14\%$ in smaller networks. We also show that SDR reaches the same accuracy that Dropout attains in 100 epochs in as few as 35 epochs.
\end{abstract}

\section{Introduction}
\label{intro}

Multi-layer neural networks have lead to remarkable performance on many kinds of benchmark tasks in text, speech and image processing.  Nonetheless, these deep layered neural networks also lead to high-dimensional, nonlinear parameter spaces that can prove difficult to search and lead to overfitting, model misspecification and poor generalization performance.   Earlier neural networks using back-propagation failed due to lack of adequate data, gradient loss recovery, and high probability of capture by poor local minima.  Deep-learning  (Hinton et al, 2006) introduced some innovations to reduce and control these overfitting and misspecification problems, including rectified linear units (ReLU), to reduce successive gradient loss and Dropout in order to avoid capture by local minima and increase generalization by effective model-averaging.  In this paper we will focus on the parameter search in the deep-layered networks despite the tsunami of data that is now available for many kinds of classification and regression tasks.     Dropout is a method that was created to mitigate the model misspecification and therefore overfitting of deep-learning applications and fundamentally to avoid poor local minima.    Specifically, Dropout implements a Bernoulli random variable with probability $p$ (“biased coin-toss”) on each update to randomly remove hidden units and their connections from the network on each update producing a sparse network architecture in which the remaining weights are updated and retained for the next Dropout step.    At the end of learning the DL network is reconstituted by calculating the expected value for each weight  $p_{w_{ij}}$ which approximates a model-averaging over an exponential set of networks.    Dropout in deep learning has been shown year after year to reduce errors on common benchmarks by more than 50\% in many cases.

In the rest of this paper we will introduce a general type of Dropout that operates at the weight level and injects gradient-dependent noise on each update called the stochastic delta rule (cf. Murray \& Andrews, 1991).    SDR redefines the scalar connection weight as  a random variable with two parameters; its mean and standard deviation.  The SDR algorithm further specifies update rules for each parameter in the random variable, which we will assume is Gaussian with parameters ($\mu_{w_{ij}}$, $\sigma_{w_{ij}}$). Note, however,  that SDR could in principle assume any random variable (gamma, beta, binomial, etc.) with at least the first two moments ( Pareto-type consequently would not be a candidate).  
	We will show that Dropout  is a special case with a binomial random variable with fixed parameters ($np$, $np(1-p)$).    Finally we will test DenseNet architectures on standard benchmarks (CIFAR-10, CIFAR-100)  with Gaussian SDR which will  show a considerable advantage over bninomial or standard Dropout.

\section{Stochastic delta rule}
\label{sdr}

It is known that actual neural transmission involves noise.   If a  cortically isolated neuron is cyclically stimulated with the exact same stimuli it will never result in the same response (Burns, et al).    Part of the motivation for SDR is based on the stochastic nature of signals transmitted through neurons in living systems.    Obviously smooth neural rate functions are based on considerable averaging over many stimulation trials.   This leads us to an implementation that suggests a synapse between two neurons could be modeled with a distribution with fixed parameters.  The possible random variables associated with such a distribution are in the time-domain likely to be a Gamma distribution (or in binned responses; see Poisson, Burns).    Here we assume a central limit theorem aggregation of i.i.d. random variables and adopt a Gaussian as a general form.  Although, there may be an advantage to longer tail distributions in the same sense that skew is required for independent component analysis (ICA).   \\

\begin{figure}[!htb]
  \centering
  \includegraphics[width=8cm]{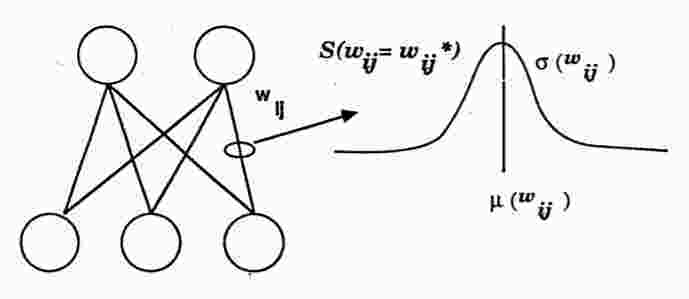}
  \caption{SDR sampling.}
\end{figure} 

At present we therefore implement the SDR algorithm with a Gaussian random variable with mean $\mu_{w_{ij}}$ and $\sigma_{w_{ij}}$ shown in Figure 1.    Each weight will  consequently be sampled from the gaussian random variable as a forward calculation.   In effect, similar to Dropout, and exponential set of networks are sampled over updates during training.   The difference at this point with Dropout is that SDR adjusts the weights and the effect of the hidden unit attached to each weight to change adapatively with the error gradient on that update.   The effect here again is similar to Dropout, except each hidden unit's response is factored across its weights (proportional to their effect on the credit assignment from the classification error). Consequently, each weight gradient itself is a random variable based on hidden unit prediction performance allowing for the system  to (1) entertain multiple response hypotheses given the same exemplar/stimulus and (2) maintain a prediction history, unlike Dropout, that is local to the hidden unit weights, is conditional on a set or even a specific exemplar and  finally (3) potentially back out of poor local minima that may result from greedy search but at the same time distant from better local minima.     The consequence of the local noise injection has a global effect on the network convergence and provides the network with greater search efficiency which we examine later on.   A final advantage, suggested by G. Hinton, was that the local noise injection may through model averaging smooth out the ravines and gullies bounded by better plauteaus in the error surface, allowing quicker and more stable convergence to better local minima.

The implementation of SDR involves three independent update rules for the parameters of the random variable representing each weight and the model averaging parameter causing the network search to eventually collapse to a single network effectively averaged over all sampled networks/exemplars/updates.   Initially a forward pass through the network involves random sampling from each weight distribution independently,
producing a $w_{ij}^*$ per connection.    This weight value is subsequently used below in the weight distribution update rules:

\begin{align*}
	S(w_{ij} = w_{ij}^{*}) = \mu_{w_{ij}} + \mu_{w_{ij}}\theta(w_{ij};0, 1)
\end{align*}

The first update rule refers to the mean of the weight distribution:

\begin{align*}
	\mu_{w_{ij}}(n + 1) = \alpha (\frac{\partial E}{\partial w_{ij}^*}) + \mu_{w_{ij}}(n)
\end{align*}

and is directly dependent on the error gradient and has learning rate $\alpha$. This is the usual delta rule update but conditioned on sample weights thus causing weight sharing through the updated mean value. The second update rule is for the standard deviation of the weight distribution (and for a Gaussian is known to be sufficient for identification).

\begin{align*}
	\sigma_{w_{ij}}(n + 1) = \beta |\frac{\partial E}{\partial w_{ij}^*}| + \sigma_{w_{ij}}(n)
\end{align*}
Again note that the standard deviation of the weight distribution is dependent on the gradient and has a multiplier coefficient of $\beta$.    And once again weight sharing is  linked through the value of the standard deviation based on the $w_{ij}^*$ sample.   A further effect of the weight standard deviation rule is that as gradient error for that mean weight value (on average) increases the hidden unit those weights connect to, is getting more uncertain and more unreliable for the network prediction.      Consequently we need one more rule to enforce the final weight average over updates.    This is another standard deviation rule (which could be combined above—however for explication purposes we have broken it out) forces the noise to “drain” out over time assuming mean and standard deviation updates are not larger then the exponential reduction of noise on each step.      This rule forces the standard deviation to converge to zero over time, causing the mean weight value to a fixed point aggregating all of the networks/updates over all samples.

\begin{align*}
	\sigma_{w_{ij}}(n + 1) = \zeta \sigma_{w_{ij}}(n + 1), \zeta < 1.
\end{align*}

Compared to standard backpropagation, Hanson (1990), showed in simple benchmark cases using parity tests that SDR would with high probability ($>.99$) converge to a solution while standard backpropagation (using 1 hidden layer) would converge less then $50$\% of the time. The scope of problems that SDR was used with often did not find a large difference if the classification was mainly linear or convex as was the case in many applications of back-propagation in the 1990s.

Next, we turn to how Dropout can be shown to be a special case of SDR.   The most obvious way to see this is to first conceive of the random search as a specific sampling distribution.

\section{Dropout as binomial fixed parameter SDR}
\label{headings}

Dropout as described before requires that hidden units per layer (except output layer) be removed in a Bernoulli process, which essentially implements  a biased coin flip at $p=0.2$, ensuring that some hidden units per layer will survive the removal leaving behind a sparser or thin network.   This process as described before also, like SDR, produces weight sharing and model averaging, reducing the effects of over-fitting.     To put the Dropout algorithm in probablistic context, consider that a Bernoulli random variable over many trials results in a Binomial distribution  with mean $np$ and standard deviation $(np(1-p))$.   The random variable is the number of removals (“successes”) over learning on the $x$ axis and the probability that the hidden unit will be removed with Binomial $(np, np(1-p))$.   If we compare Dropout to SDR in the same network, the difference we observe is in terms  of whether the random process is affecting weights or hidden units.   In Figure 3, we illustrate the convergence of Dropout as hidden unit Binomial sampling.   It can readily be seen that the key difference between the two is that SDR adaptively updates the random variable parameters for subsequent sampling and Dropout samples from  a Binomial random variable with fixed  parameters (mean, standard deviation at $p$).   One other critical  difference is that the weight sharing in SDR is more local per hidden unit than that of Dropout, but is essentially the same coupling over sampling trials with Dropout creating an equivalence class per hidden unit and thus creating a coarser network history. \\

\begin{figure}[!htb]
  \centering
  \includegraphics[width=8cm]{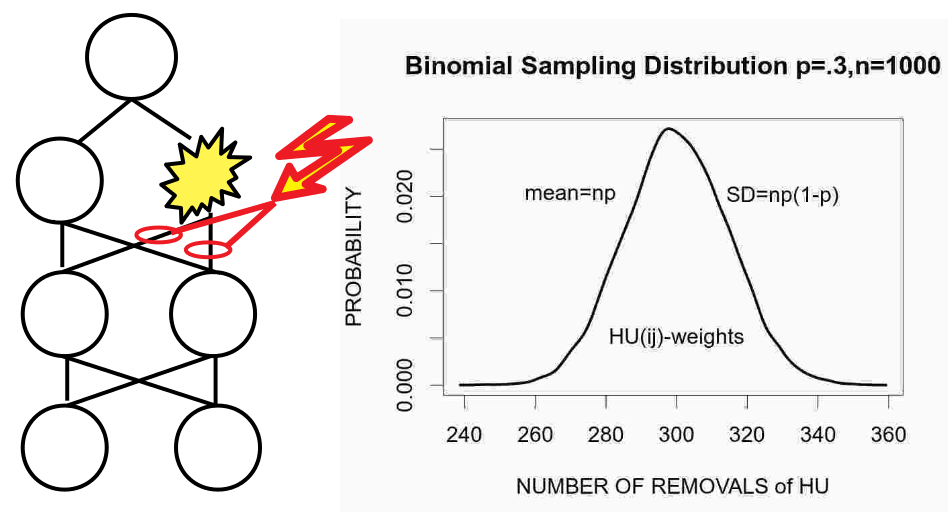}
  \caption{Dropout sampling.}
\end{figure} 

Showing that SDR is a generalized form of Dropout opens the door for many kinds of variations in random search, that would be potentially more directed and efficient than the fixed parameter search that Dropout represents.   It is possible that longer tailed distribution for example (Gamma, Beta, LogNormal..etc) would be more similar to the distributions underlying neural spikes trains, thus allowing us to provide further connections to the stochastic nature of neural transmission.   More critically in this context, does the increase in parameters that SDR represents provide more efficient and robust search that would increase performance in classification domains already well tested with many kinds of variations of deep learning with Dropout?

In what follows we implement and test a state of the art deep neural network, in this case DenseNet (Huang, 2017) with standard benchmark image tests (CFAR-10, CIFAR-100).   Here we intend to show paired tests with PyTorch implementations holding the learning parameters (except for the random search algorithms—SDR or Dropout) constant over various conditions.

\section{Implementation}

Tests were conducted on a compute server with two Intel Xeon E5-2650 CPUs, 512GB RAM, and two NVIDIA P100 GPUs. We used a modified DenseNet model implemented in PyTorch, originally by A. Veit (2017). The model uses a DenseNet-40, DenseNet-100, and DenseNet-BC 250 network trained on CIFAR-10 and CIFAR-100, with a growth rate of $k=12$, batch size of $100$, and $100$ epochs, with the other parameters being the same as the original DenseNet implementation. The learning rate drop ratio is maintained as well, with $\alpha$/LR dropping at 50\% and 75\% of the run. Further work is being done to extend the variety in parameters and datasets (similar to the experiments done with the original DenseNet implementation). The model without SDR used a Dropout rate of $0.2$, i.e a $20$\% chance that each neuron is dropped out. The SDR implementation used parameters that varied according to the size of the network and the number of classes, but the values were generally around $\alpha=0.25, \beta=0.05, \zeta=0.7$. We hyperbolically annealed $\zeta$ for smaller networks and exponentially annealed it for larger networks so as to reduce the influence of the standard deviations as the model converges. The standard deviations were initialized using a halved Xavier initialization and were updated twice every epoch, in the middle and at the end, for DenseNet-BC 250 and DenseNet-100 and after every batch for the others (the number of updates per epoch has an effect on the overall performance and can be treated as a hyperparameter. We noticed that larger networks needed fewer gradient updates than smaller networks). The propagation of $\zeta$ is split between the earlier layers and the deeper layers, with the $\zeta$ value in the earlier layers being $90\%$ of the specified value. The code used for implementing and testing SDR is publicly available\footnote{\url{https://github.com/noahfl/sdr-densenet-pytorch}}.

\section{Results}

\subsection{Performance Benchmarks}

\begin{figure}[!htb]
  \centering
  \includegraphics[width=8cm]{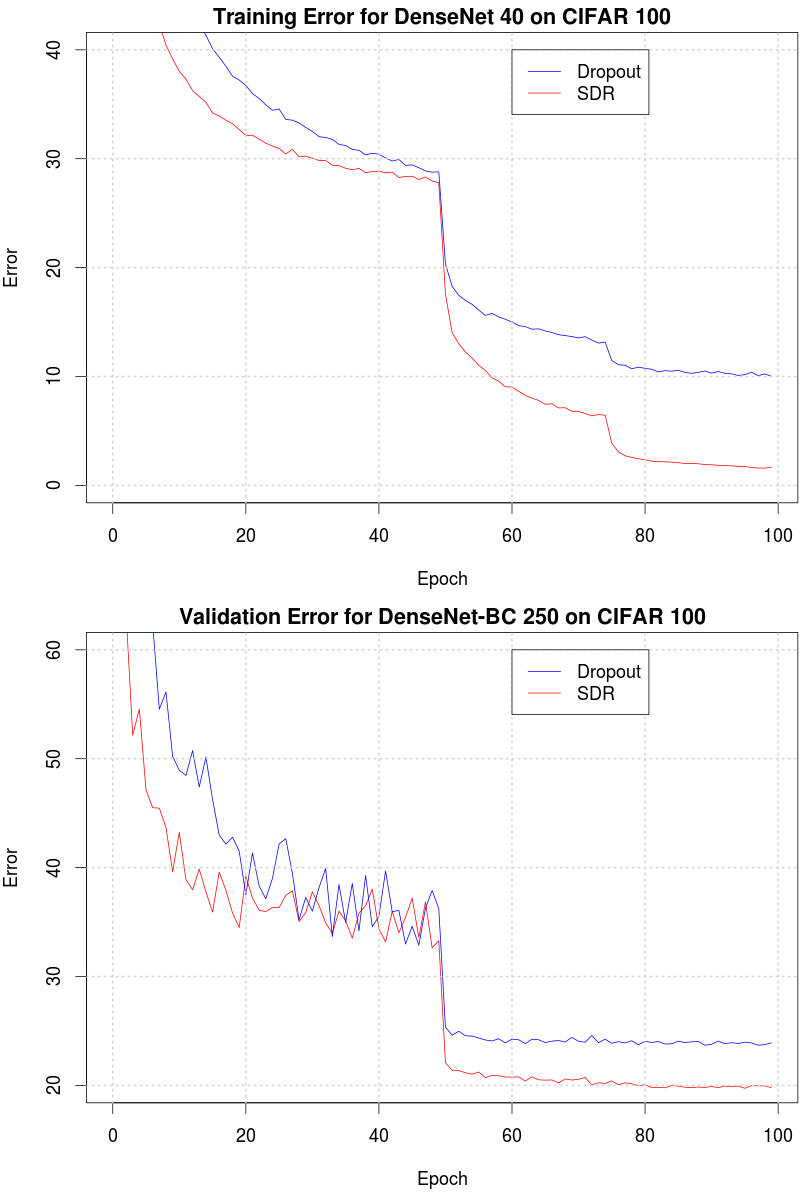}
  \caption{Error plots for DenseNet-40 training error and DenseNet-BC 250 validation error. Training error for DenseNet-40 is reduced by 83\% and the validation error for DenseNet-BC 250 is reduced by 17\%.}
\end{figure}

\begin{table}[!htb]
  \caption{Top-1 error validation rates at end of training of DenseNet-SDR compared to DenseNet with Dropout.}
  \centering
  \begin{tabular}{lll}
    \toprule
    \multicolumn{3}{c}{Dataset}                  \\
    \cmidrule(r){2-3}
    Model     & CIFAR-10     & CIFAR-100 \\
    \midrule
    DenseNet-40 & $6.88$  & $27.88$     \\
    (k=12) \\
    DenseNet-100 & -  & $24.67$     \\
    (k=12) \\
    DenseNet-BC 250 & -  & $23.91$     \\
    (k=12) \\
    \midrule
    DenseNet-40 with SDR     & \textbf{5.91} & \textbf{24.58}      \\
    (k=12) \\
    DenseNet-100 with SDR     & - & \textbf{21.72}      \\
    (k=12) \\
    DenseNet-BC 250 with SDR     & - & \textbf{19.79}      \\
    (k=12)    \\ 
    
    \bottomrule
  \end{tabular}
\end{table}

These results show that replacing Dropout with SDR in DenseNet tests yields approx. $12-14\%$ reduction in error smaller networks, and $17\%$ reduction in error on DenseNet-BC 250 on CIFAR 100. Error results for the DenseNet implementation with Dropout are slightly lower than in the original DenseNet paper as it was observed that using a larger batch size resulted in higher overall accuracy.

\begin{table}[!htb]
  \caption{Training losses of DenseNet-SDR compared to DenseNet with Dropout at end of training.}
  \centering
  \begin{tabular}{lll}
    \toprule
    \multicolumn{3}{c}{Dataset}                  \\
    \cmidrule(r){2-3}
    Model     & CIFAR-10     & CIFAR-100 \\
    \midrule
    DenseNet-40 & $1.85$  & $10.01$     \\
    (k=12) \\
    DenseNet-100 & -  & $1.17$     \\
    (k=12) \\
    DenseNet-BC 250 & -  & $1.24$     \\
    (k=12) \\
    \midrule
    DenseNet-40 with SDR     & \textbf{0.24} & \textbf{0.89}      \\
    (k=12) \\
    DenseNet-100 with SDR     & - & \textbf{0.15}      \\
    (k=12) \\
    DenseNet-BC 250 with SDR     & - & \textbf{0.11}      \\
    (k=12)    \\ 
    
    \bottomrule
  \end{tabular}
\end{table}

Tests were conducted to determine improvements in training error across benchmarks. As shown in Table 2, SDR shows an $80\%$+ reduction in training error across all benchmarks, with the parameters used shown in the code repository. This reduction may be applicable to other areas of deep learning, such as generative adversarial networks, where encoding of the training set is crucial to generative performance.


\newpage

\subsection{Speed increases}

\begin{figure}[!htb]
  \centering
  \includegraphics[width=8cm]{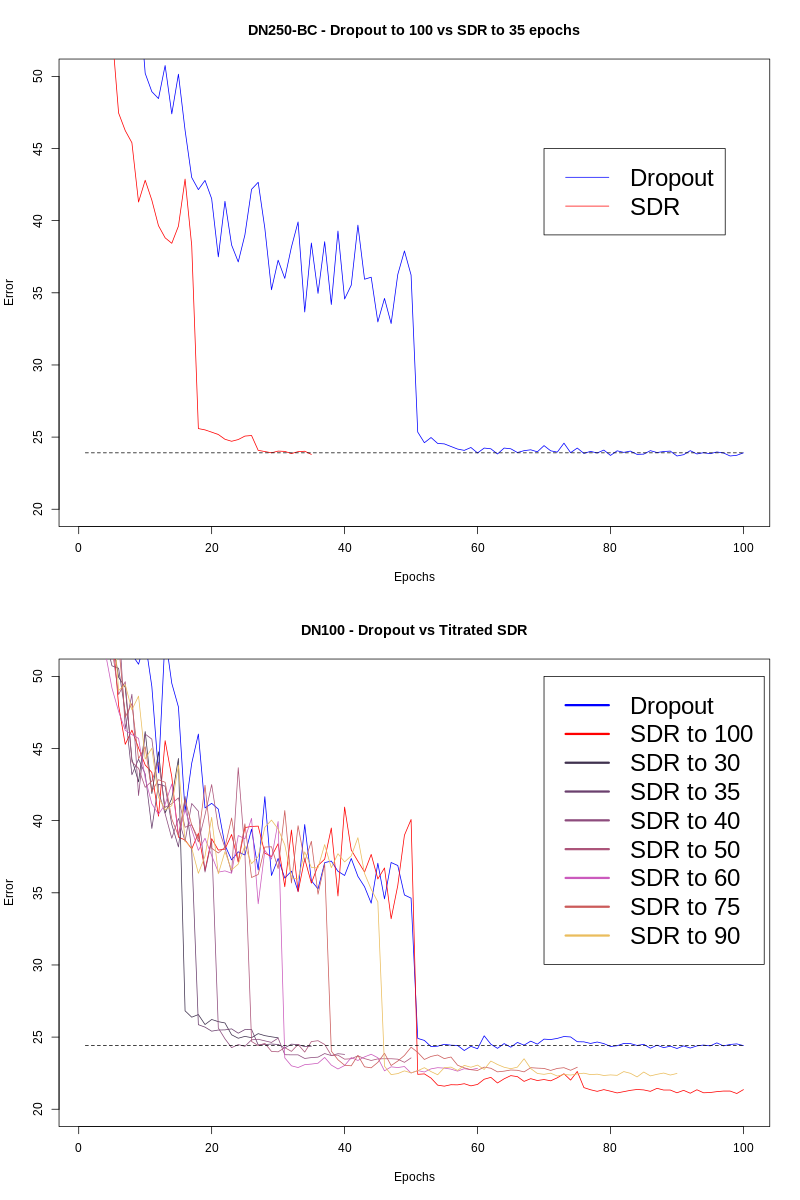}
  \caption{(Top) Error plots showing DN250-BC on CIFAR 100 with Dropout run out to 100 epochs vs. DN250-BC with SDR run to 35 epochs. (Bottom) DN100 with Dropout on CIFAR 100 vs. SDR, titrated from 100 epochs down to 30. In both images the dotted line shows the final error reached by Dropout.}
\end{figure}

Separate tests were conducted in order to find any increases in training speed to the training error reached by Dropout at the end of 100 epochs. We titrated the number of epochs, while maintaining the ratio of $\alpha$ drops (e.g running for 40 epochs would yield $\alpha$ drops at epochs 20 and 30) in order to determine the minimum number of epochs required to reach Dropout's final training error. Results are shown in the table below.

\begin{table}[!htb]
  \caption{Number of epochs required for each SDR model to reach the respective final validation error for Dropout. }
  \centering
  \begin{tabular}{lll}
    \toprule
    \multicolumn{3}{c}{Dataset}                  \\
    \cmidrule(r){2-3}
    Model     & CIFAR-10     & CIFAR-100 \\
    \midrule
    DenseNet-40 with SDR     & \textbf{45} & \textbf{35}      \\
    (k=12) \\
    DenseNet-100 with SDR     & - & \textbf{35}      \\
    (k=12) \\
    DenseNet-BC 250 with SDR     & - & \textbf{35}      \\
    (k=12)    \\ 
    
    \bottomrule
  \end{tabular}
\end{table}

As shown above, SDR allows for the same level of accuracy given by Dropout in as little as 35\% training epochs to Dropout's 100. \\

Preliminary results on ImageNet using a 121-layer DenseNet model show approx. $9\%$ improvement over dropout (error $\textit{28.35}$ with dropout and error $\textbf{25.80}$ with SDR). Further results on ImageNet are being produced and will be added as they are generated.

\newpage
$ $ 
\newpage

\section{Discussion}

\begin{figure}[!htb]
  \centering
  \includegraphics[width=8cm]{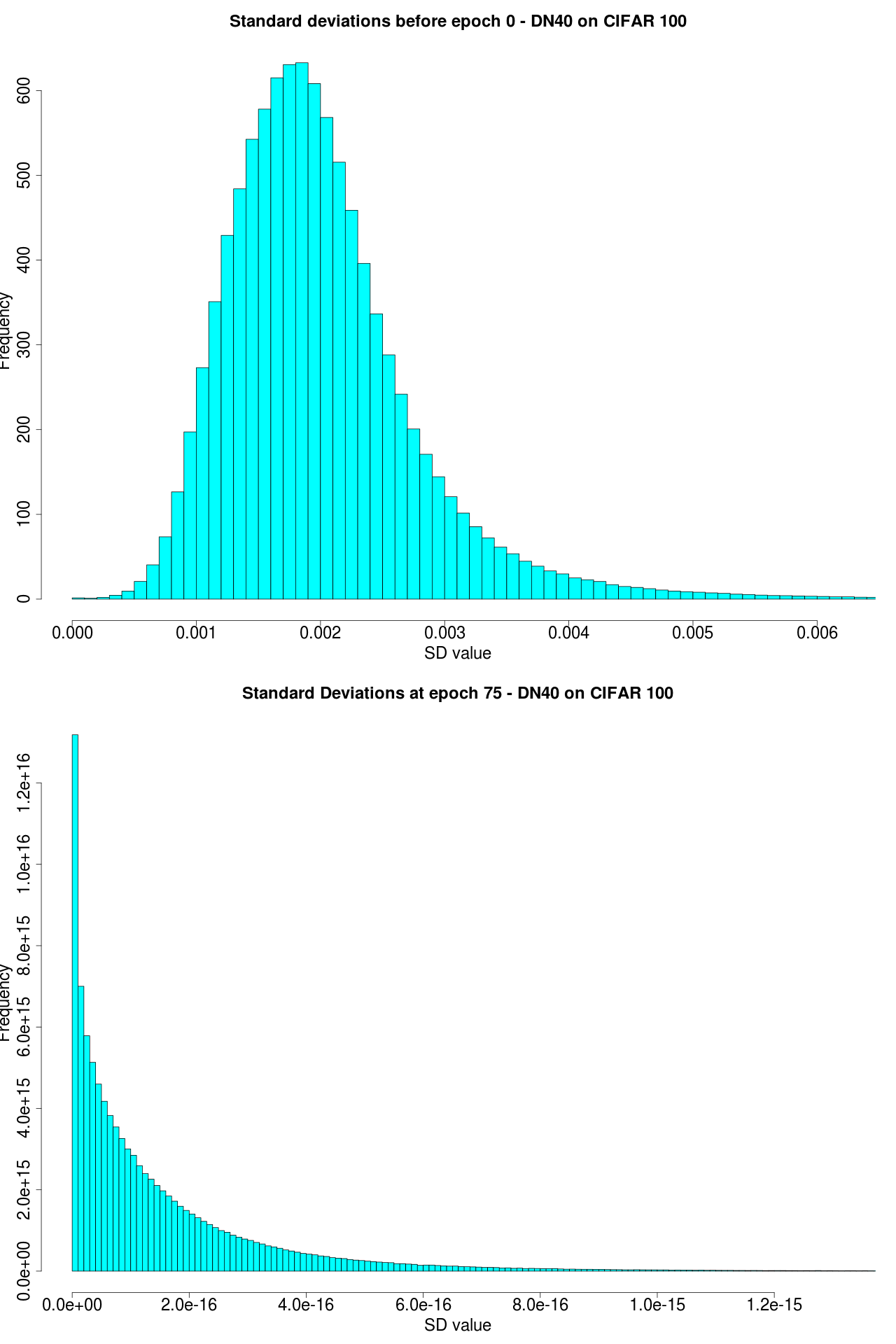}
  \caption{(Above) Initialized values of the standard deviations prior to training. (Below) values of standard deviations after 75 training epochs. Note the axes on both figures and how the standard deviations have converged nearly to zero towards the end of training.}
\end{figure} 

In this paper we have shown how a basic machine learning algorithm (Dropout) that implements stochastic search and helps prevent over-fitting is a special case of an older algorithm, the Stochastic Delta Rule, which is based on a Gaussian random sampling on weights and adaptable random variable parameters (in this case a mean value, $\mu_{w_{ij}}$,  and a standard deviation, $\sigma_{w_{ij}}$). Further, we were able to show how SDR outperforms Dropout in a state of the art DL classifier on standard benchmarks, showing improvements in test error by approx. $10$+\% in smaller networks and approx. $17\%$ in larger networks and improvements in training error of $80$+\%. In addition better encoding training data, SDR better generalizes to testing data. From an implementation standpoint, it is straightforward to implement and can be written in approximately 30 lines of code. All that is required is access to the gradients and to the weights; SDR can be inserted into virtually any training implementation with this access. SDR opens up a novel set of directions for deep learning search methods which include various random variable selections that may reflect more biophysical details of neural noise, or provide more parameters to code the prediction history within the network models, thus increasing the efficiency and efficacy of the underlying search process.

\section*{References}

\small

[1] Baldi, P. and PJ. Sadowski, Understanding Dropout, In Advances in Neural Information Processing Systems 26 ppg:2814:2822, 2013

[2] Burns, B. Delisle. The Uncertain Nervous System. Edward Arnold Publ., 1968. 

[3] Hanson, S. J. "A Stochastic Version of the Delta Rule", Physica D, vol. 42, pp. 265-272, 1990. 

[4] Hinton, G. E. and Salakhutdinov, R. R  Reducing the dimensionality of data with neural networks.
Science, Vol. 313. no. 5786, pp. 504 - 507, 28 July 2006.

[5] Huang G. ,Liu, Z.  Weinberger, K, \& Maaten, L. (2017) Densely Connected Convolutional Networks. arXiv:1608.06993

[6] Khlyestov, I. DenseNet with TensorFlow. (2017) GitHub repository. https://github.com/ikhlestov/vision\_networks

[7] Murray, A. F., "Analog Noise-Enhanced Learning in Neural Network Circuits," Electronics Letters, vol. 2, no. 17, pp. 1546-1548, 1991. 

[8] Srivastava, N., Hinton, G., Krizhevsky, A., Sutskever, I., and Salakhutdinov, R. R. Improving neural networks by preventing co-adaptation of feature detectors. http://arxiv.org/abs/1207.0580, 2012.

[9] Veit, A. A PyTorch Implementation for Densely Connected Convolutional Networks (DenseNets). (2017) GitHub repository. https://github.com/andreasveit/densenet-pytorch/

\end{document}